\pdfoutput=1

\documentclass[11pt]{article}

\usepackage{acl}

\usepackage{times}
\usepackage{latexsym}

\usepackage[T1]{fontenc}

\usepackage[utf8]{inputenc}

\usepackage{microtype}

\usepackage{booktabs}
\usepackage{xcolor}
\usepackage{mathtools}
\usepackage{bm}
\newcommand{\bx}{\bm{x}}
\usepackage{amsmath}
\usepackage{amssymb}
\newcommand{\E}{\mathop{\mathbb{E}}}
\usepackage{multirow}

\newcommand{\eg}{\textit{e}.\textit{g}. }

\title{DATE: Detecting Anomalies in Text via Self-Supervision of Transformers}

\author{Andrei Manolache$^{1, 2}$ \and Florin Brad$^{1}$ \and Elena Burceanu$^{1, 2, 3}$ \\
        $^1$Bitdefender\\
        $^2$University of Bucharest, Romania\\
        $^3$Institute of Mathematics of the Romanian Academy\\
        \texttt{\{amanolache, fbrad, eburceanu\}@bitdefender.com}\\ }

\begin{document}
\maketitle

\begin{abstract}
Leveraging deep learning models for Anomaly Detection (AD) has seen widespread use in recent years due to superior performances over traditional methods. Recent deep methods for anomalies in images learn better features of normality in an end-to-end self-supervised setting. These methods train a model to discriminate between different transformations applied to visual data and then use the output to compute an anomaly score. We use this approach for AD in text, by introducing a novel pretext task on text sequences. We learn our DATE model end-to-end, enforcing two independent and complementary self-supervision signals, one at the token-level and one at the sequence-level. Under this new task formulation, we show strong quantitative and qualitative results on the 20Newsgroups and AG News datasets. In the \emph{semi-supervised} setting, we outperform state-of-the-art results by +13.5\% and +6.9\%, respectively (AUROC). In the \emph{unsupervised} configuration, DATE surpasses all other methods even when 10\% of its training data is contaminated with outliers (compared with 0\% for the others).

\end{abstract}

\section{Introduction}
Anomaly Detection (AD) can be intuitively defined as the task of identifying examples that deviate from the other ones to a degree that arouses suspicion \cite{Hawkins1980}. Research into AD spans several decades \cite{Chandola2009,Aggarwal2015} and has proved fruitful in several real-world problems, such as intrusion detection systems \cite{Banoth2017}, credit card fraud detection \cite{Dorronsoro1997}, and manufacturing \cite{Kammerer2019}.

Our DATE method is applicable in the \emph{semi-supervised} AD setting, in which we only train on clean, labeled normal examples, as well as the \emph{unsupervised} AD setting, where both unlabeled normal and abnormal data are used for training. Typical deep learning approaches in AD involve learning features of normality using autoencoders \cite{Hawkins2002,Sakurada2014,Chen2017} or generative adversarial networks \cite{Schlegl2017}. Under this setup, anomalous examples lead to a higher reconstruction error or differ significantly compared with generated samples.

Recent deep AD methods for images learn more effective features of visual normality through \emph{self-supervision}, by training a deep neural network to discriminate between different transformations applied to the input images \cite{Golan2018,neurips2019}. An anomaly score is then computed by aggregating model predictions over several transformed input samples.

\begin{figure}[t!]
\centering
\includegraphics[width=0.99\columnwidth]{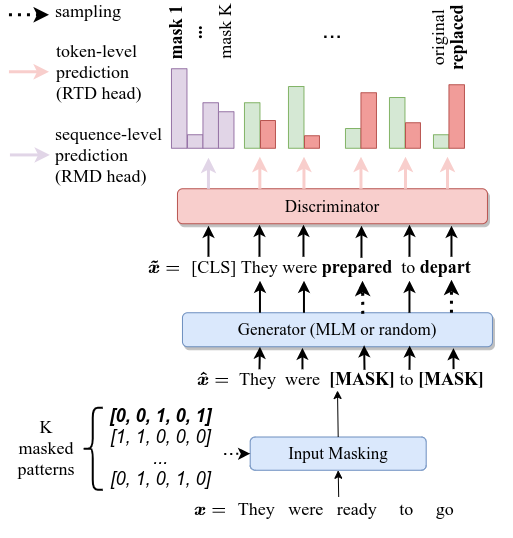}
\caption{DATE Training. Firstly, the input sequence is masked using a sampled masked pattern and a generator fills in new tokens in place of the masked ones. Secondly, the discriminator receives supervision signals from two tasks: RMD (which mask pattern was applied to the input sequence) and RTD (the per-token status: original or replaced).}
\label{fig:date_train}
\end{figure}

\begin{figure}[t!]
\centering
\includegraphics[width=0.8\columnwidth]{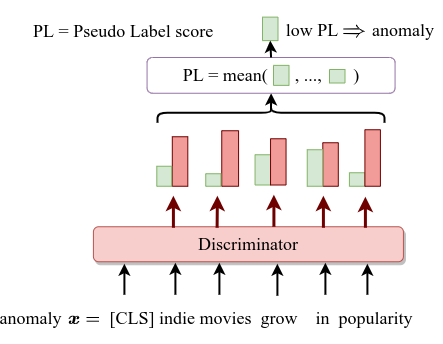}
\caption{DATE Testing. The input text sequence is fed to the discriminator, resulting in token-level probabilities for the normal class, which are further aggregated into an anomaly score, as detailed in Sec.\ref{sec:pl_score}. For deciding whether a sample is either normal or abnormal, we aggregate over all of its tokens.}
\label{fig:date_test}
\end{figure}

We adapt those self-supervised classification methods for AD from vision to learn anomaly scores indicative of text normality. ELECTRA~\cite{Clark2020} proposes an efficient language representation learner, which solves the \emph{Replaced Token Detection} (RTD) task. Here the input tokens are plausibly corrupted with a BERT-based~\cite{Devlin2018} generator, and then a discriminator predicts for each token if it is real or replaced by the generator. In a similar manner, we introduce a complementary sequence-level pretext task called \emph{Replaced Mask Detection} (RMD), where we enforce the discriminator to predict the predefined \emph{mask pattern} used when choosing what tokens to replace. For instance, given the input text \emph{`They were ready to go`} and the mask pattern \texttt{[0, 0, 1, 0, 1]}, the corrupted text could be \emph{`They were prepared to advance`}. The RMD multi-class classification task asks which \emph{mask pattern} (out of K such patterns) was used to corrupt the original text, based on the corrupted text. Our generator-discriminator model solves both the RMD and the RTD task and then computes the anomaly scores based on the output probabilities, as visually explained in detail Fig.~\ref{fig:date_train}-\ref{fig:date_test}.

We notably simplify the computation of the Pseudo Label (PL) anomaly score~\cite{neurips2019} by removing the dependency on running over multiple transformations and enabling it to work with token-level predictions. This significantly speeds up the PL score evaluation.

To our knowledge, DATE is the first end-to-end deep AD method on text that uses self-supervised classification models to produce normality scores. Our \textbf{contributions} are summarized below:

\begin{itemize}
    \item We introduce a sequence-level self-supervised task called \emph{Replaced Mask Detection} to distinguish between different transformations applied to a text. Jointly optimizing both sequence and token-level tasks stabilizes training, improving the AD performance.
    \item We compute an efficient Pseudo Label score for anomalies, by removing the need for evaluating multiple transformations, allowing it to work directly on individual tokens probabilities. This makes our model faster and its results more interpretable.
    \item We outperform existing state-of-the-art semi-supervised AD methods on text by a large margin (AUROC) on two datasets: 20Newsgroups (+13.5\%) and AG News (+6.9\%). Moreover, in unsupervised AD settings, even with 10\% outliers in training data, DATE surpasses all other methods trained with 0\% outliers.
\end{itemize}

\section{Related Work}
Our work relates to self-supervision for language representation as well as self-supervision for learning features of normality in AD.

\subsection{Self-supervision for NLP}
Self-supervision has been the bedrock of learning good feature representations in NLP. The earliest neural methods leveraged shallow models to produce static word embeddings such as word2vec~\cite{Mikolov2013}, GloVe~\cite{Pennington2014} or fastText~\cite{Bojanowski2017,Joulin2017}. More recently, contextual word embeddings have produced state-of-the-art results in many NLP tasks, enabled by Transformer-based~\cite{Vaswani2017} or LSTM-based~\cite{Hochreiter1997} architectures, trained with language modeling~\cite{Peters2018,Radford2019} or masked language modeling~\cite{Devlin2018} tasks.

Many improvements and adaptations have been proposed over the original BERT, which address other languages \cite{Martin2020,Vries2019}, domain specific solutions \cite{Beltagy2019,Lee2020} or more efficient pre-training models such as ALBERT~\cite{Lan2019} or ELECTRA~\cite{Clark2020}. 
ELECTRA pre-trains a BERT-like generator and discriminator with a Replacement Token Detection (RTD) Task. The generator substitutes masked tokens with likely alternatives and the discriminator is trained to distinguish between the original and masked tokens.

\subsection{Self-supervised classification for AD}
Typical representation learning approaches to deep AD involve learning features of normality using autoencoders~\cite{Hawkins2002, Sakurada2014,Chen2017} or generative adversarial networks~\cite{Schlegl2017}. More recent methods train the discriminator in a self-supervised fashion, leading to better normality features and anomaly scores. These solutions mostly focus on image data~\cite{Golan2018, neurips2019} and train a model to distinguish between different transformations applied to the images (\eg rotation, flipping, shifting). An interesting property that justifies self-supervision under unsupervised AD is called \emph{inlier priority}~\cite{neurips2019}, which states that during training, inliers (normal instances) induce higher gradient magnitudes than outliers, biasing the network's update directions towards reducing their loss. Due to this property, the outputs for \emph{inliers} are more consistent than for \emph{outliers}, enabling them to be used as anomaly scores.

\subsection{AD for text}
There are a few shallow methods for AD on text, usually operating on traditional document-term matrices. One of them uses one-class SVMs~\cite{Schlkopf2001} over different sparse document representations~\cite{Manevitz2001OneClassSF}. Another method uses non-negative matrix factorization to decompose the term-document matrix into a low-rank and an outlier matrix~\cite{Kannan2017}. LDA-based~\cite{Blei2003} clustering algorithms are augmented with semantic context derived from WordNet~\cite{Miller1995} or from the web to detect anomalies~\cite{Mahapatra2012}.

\subsection{Deep AD for text}
While many deep AD methods have been developed for other domains, few approaches use neural networks or pre-trained word embeddings for text anomalies. Earlier methods use autoencoders~\cite{Manevitz2007} to build document representations. More recently, pre-trained word embeddings and self-attention were used to build contextual word embeddings~\cite{acl2019}. These are jointly optimized with a set of \emph{context vectors}, which act as topic centroids. The network thus discovers relevant topics and transforms normal examples such that their contextual word embeddings stay close to the topic centroids. Under this setup, anomalous instances have contextual word embeddings which on average deviate more from the centroids.

\section{Our Approach}
Our method is called \textbf{DATE} for 'Detecting Anomalies in Text using ELECTRA'. We propose an end-to-end AD approach for the discrete text domain that combines our novel self-supervised task (Replaced Mask Detection), a powerful representation learner for text (ELECTRA), and an AD score tailored for sequential data. We present next the components of our model and a visual representation for the training and testing pipeline in Fig.~\ref{fig:date_train}-\ref{fig:date_test}.

\subsection{Replaced Mask Detection task}
We introduce a novel self-supervised task for text, called Replaced Mask Detection (RMD). This discriminative task creates training data by transforming an existing text using one out of K given operations. It further asks to predict the correct operation, given the transformed text. The transformation over the text consists of two steps: 1) \emph{masking} some of the input words using a predefined mask pattern and 2) replacing the masked words with alternative ones (\eg 'car' with 'taxi').

\paragraph{Input masking.} Let $\bm{m} \in \{0,1\}^T$ be a mask pattern corresponding to the text input $\bx~=~[x_1, x_2, ..., x_T]$. For training, we generate and fix $K$ mask patterns $\bm{m}^{(1)}, \bm{m}^{(2)}, ..., \bm{m}^{(K)}$ by randomly sampling a constant number of ones. Instead of masking random tokens on-the-fly as in ELECTRA, we first sample a mask pattern from the K predefined ones. Next we apply it to the input, as in Fig.~\ref{fig:date_train}.
Let $\bm{\hat{x}}(\bm{m})=[\hat{x}_1, \hat{x}_2, ..., \hat{x}_T]$ be the input sequence $\bx$, masked with $\bm{m}$, where:
\begin{equation*}
    \hat{x}_i=
    \begin{cases}
      x_i, & \bm{m}_i=0 \\
      \texttt{[MASK]}, & \bm{m}_i=1
    \end{cases}
\end{equation*}
For instance, given an input $\bx=$~[bank, hikes, prices, before, election] and a mask pattern $\bm{m} = [0, 0, 1, 0, 1]$, the masked input is $\bm{\hat{x}}(\bm{m})=$~[bank, hikes, \texttt{[MASK]}, before, \texttt{[MASK]}].

\paragraph{Replacing [MASK]s.} Each masked token can be replaced with a word token (\eg by sampling uniformly from the vocabulary). For more plausible alternatives, masked tokens can be sampled from a Masked Language Model (MLM) generator such as BERT, which outputs a probability distribution $P_G$ over the vocabulary, for each token. Let $\widetilde{\bx}(\bm{m})=[\widetilde{x}_1, \widetilde{x}_2, ..., \widetilde{x}_T]$ be the plausibly corrupted text, where:
\begin{equation*}
    \widetilde{x_i}=
    \begin{cases}
      x_i, & \bm{m}_i=0 \\
      w_i\sim P_{G}(x_i|\bm{\hat{x}}(\bm{m});\theta_G), & \bm{m}_i=1
    \end{cases}
\end{equation*}
For instance, given the masked input $\bm{\hat{x}}(\bm{m})=$~[bank, hikes, \texttt{[MASK]}, before, \texttt{[MASK]}], a plausibly corrupted input is $\widetilde{\bx}(\bm{m})=$~[bank, hikes, fees, before, referendum].

\paragraph{Connecting RMD and RTD tasks.} RTD is a binary sequence tagging task, where some tokens in the input are corrupted with plausible alternatives, similarly to RMD. The discriminator must then predict for each token if it's the \emph{original} token or a \emph{replaced} one. Distinctly from RTD, which is a \emph{token-level} discriminative task, RMD is a \emph{sequence-level} one, where the model distinguishes between a fixed number of predefined transformations applied to the input. As such, RMD can be seen as the text counterpart task for the self-supervised classification of geometric alterations applied to images~\cite{Golan2018,neurips2019}. While RTD predictions could be used to sequentially predict an entire mask pattern, they can lead to masks that are not part of the predefined K patterns. But the RMD constraint overcomes this behaviour. We thus train DATE to solve both tasks simultaneously, which increases the AD performance compared to solving one task only, as shown in Sec.~\ref{sec: ex-ablation}. Furthermore, this approach also improves training stability.

\subsection{DATE Architecture}
We solve RMD and RTD by jointly training a generator, G, and a discriminator, D. G is an MLM used to \emph{replace} the masked tokens with plausible alternatives. We also consider a setup with a \emph{random generator}, in which we sample tokens uniformly from the vocabulary. D is a deep neural network with two prediction heads used to distinguish between \emph{corrupted} and original tokens (RTD) and to predict which mask pattern was applied to the corrupted input (RMD). At test time, G is discarded and D's probabilities are used to compute an anomaly score.

Both G and D models are based on a BERT encoder, which consists of several stacked Transformer blocks~\cite{Vaswani2017}. The BERT encoder transforms an input token sequence $\bx~=~[x_1, x_2, ..., x_T]$ into a sequence of contextualized word embeddings $h(\bx) = [h_1, h_2, ..., h_T]$.

\paragraph{Generator.} G is a BERT encoder with a linear layer on top that outputs the probability distribution $P_G$ for each token. The generator is trained using the MLM loss:
\begin{equation}
    \mathcal{L}_{MLM}= E\biggl[\sum_{\substack{i=1;\\s.t. m_i=1}}^{T} -\log P_G(x_i|\bm{\hat{x}}(\bm{m});\theta_G)\biggr]
\end{equation}

\paragraph{Discriminator.} D is a BERT encoder with two prediction heads applied over the contextualized word representations:

\noindent \textbf{i. RMD head.} This head outputs a vector of logits for all mask patterns $\bm{o}=[o_1,...,o_K]$. We use the contextualized hidden vector $h_{\texttt{[CLS]}}$ (corresponding to the $\texttt{[CLS]}$ special token at the beginning of the input) for computing the mask logits $\bm{o}$ and $P_M$, the probability of each mask pattern:
\begin{equation}
    P_{M}(\bm{m}=\bm{m}^{(k)}|\widetilde{\bx}(\bm{m}^{(k)}); \theta_D)=\frac{exp(o_k)}{\sum_{i=1}^{K} exp(o_i)}
\end{equation}

\noindent \textbf{ii. RTD head.} This head outputs scores for the two classes (\emph{original} and \emph{replaced}) for each token $x_1, x_2, ..., x_T$, by using the contextualized hidden vectors $h_1, h_2, ..., h_T$.

\paragraph{Loss.} We train the DATE network in a maximum-likelihood fashion using the $\mathcal{L}_{DATE}$ loss:
\begin{equation}
     \min_{\theta_D, \theta_G}  \sum_{\bx \in \mathcal{X}} \mathcal{L}_{DATE}(\theta_D,\theta_G;\bx)
\end{equation}

\noindent The loss contains both the token-level losses in ELECTRA, as well as the sequence-level mask detection loss $\mathcal{L}_{RMD}$:
\begin{align}
\begin{split}
    \mathcal{L}_{DATE}(\theta_D,\theta_G;\bx) = \mu\mathcal{L}_{RMD}(\theta_D;\bx) +
    \\\mathcal{L}_{MLM}(\theta_G;\bx) + \lambda\mathcal{L}_{RTD}(\theta_D;\bx),
\end{split}
\end{align}
where the discriminator losses are:
\begin{align}
    \mathcal{L}_{RMD}&=\E \biggl[-\log P_{M}(\bm{m}|\widetilde{\bx}(\bm{m});\theta_D)\biggr], \\
    \mathcal{L}_{RTD}&=\E\biggl[ \sum_{\substack{i=1;\\x_i\ne \texttt{[CLS]}}}^{T} -\log P_D(m_i|\bm{\widetilde{\bx}}(\bm{m});\theta_D)\biggr],
\end{align}
where $P_D$ is the probability distribution that a token was replaced or not.

The ELECTRA loss enables D to learn good feature representations for language understanding. Our RMD loss puts the representation in a larger sequence-level context. After pre-training, G is discarded and D can be used as a general-purpose text encoder for downstream tasks. Output probabilities from D are further used to compute an anomaly score for new examples.

\subsection{Anomaly Detection score}
\label{sec:pl_score}
We adapt the Pseudo Label (PL) based score from the \emph{$E^{3}Outlier$} framework \cite{neurips2019} in a novel and efficient way. In its general form, the PL score aggregates responses corresponding to multiple transformations of $\bx$. This approach requires $k$ input transformations over an input $\bx$ and $k$ forward passes through a discriminator. It then takes the probability of the ground truth transformation and averages it over all $k$ transformations. 

To compute PL for our RMD task, we take $\bx$ to be our input text and the K mask patterns as the possible transformations. We corrupt $\bx$ with mask $\bm{m}^{(i)}$ and feed the resulted text to the discriminator. We take the probability of the i-th mask from the RMD head. We repeat this process $k$ times and average over the probabilities of the correct mask pattern. This formulation requires $k$ feed-forward steps through the DATE network, which slows down inference. We propose a more computationally efficient approach next.




\paragraph{PL over RTD classification scores.} 
Instead of aggregating \emph{sequence-level} responses from multiple transformations over the input, we can aggregate \emph{token-level} responses from a single model over the input to compute an anomaly score. More specifically, we can discard the generator and feed the original input text to the discriminator directly. We then use the probability of each token being \emph{original} (not \emph{corrupted}) and then average over all the tokens in the sequence:
\begin{equation}
    \label{eq: PL_score}
    PL_{RTD}(x) = \frac{1}{T} \sum_{i=1}^T
    P_D(m_i=0|\bm{\widetilde{\bx}}(\bm{m}^{(0)});\theta_D),
\end{equation}
where $\bm{m}^{(0)}=[0, 0, ..., 0]$ effectively leaves the input unchanged.
As can be seen in Fig.~\ref{fig:date_test}, the RTD head will be less certain in predicting the \emph{original} class for \emph{outliers} (having a probability distribution unseen at training time), which will lead to lower PL scores for \emph{outliers} and higher PL scores for \emph{inliers}. We use PL at testing time, when the entire input is either normal or abnormal. Our method also speeds up inference, since we only do one feed-forward pass through the discriminator instead of $k$ passes. Moreover, having a per token anomaly score helps us better understand and visualize the behavior of our model, as shown in Fig.~\ref{fig: qual_ex}.

\section{Experimental analysis}
In this section, we detail the empirical validation of our method by presenting: the semi-supervised and unsupervised experimental setup, a comprehensive ablation study on DATE, and the comparison with state-of-the-art on the semi-supervised and unsupervised AD tasks. DATE does not use any form of pre-training or knowledge transfer (from other datasets or tasks), learning all the embeddings from scratch. Using pre-training would introduce unwanted prior knowledge about the outliers, making our model considering them known (normal).

\subsection{Experimental setup}
We describe next the Anomaly Detection setup, the datasets and the implementation details of our model. We make the code publicly available~\footnote{\url{https://github.com/bit-ml/date}}.

\paragraph{Anomaly Detection setup.} We use a semi-supervised setting in Sec.~\ref{sec: ex-ablation}-\ref{sec: ex-ssad} and an unsupervised one in Sec.~\ref{sec: ex-uad}. In the semi-supervised case, we successively treat one class as normal (\emph{inliers}) and all the other classes as abnormal (\emph{outliers}). In the unsupervised AD setting, we add a fraction of outliers to the inliers training set, thus contaminating it. We compute the Area Under the Receiver Operating Curve (AUROC) for comparing our method with the previous state-of-the-art. For a better understanding of our model's performance in an unbalanced dataset, we report the Area Under the Precision-Recall curve (AUPR) for inliers and outliers per split in the supplementary material~\ref{apx:prauc_experiments}.

\paragraph{Datasets.} We test our solution using two text classification datasets, after stripping headers and other metadata. For the first dataset, 20Newsgroups, we keep the exact setup, splits, and preprocessing (lowercase, removal of: punctuation, number, stop word and short words) as in \cite{acl2019}, ensuring a fair comparison with previous text anomaly detection methods. As for the second dataset, we use a significantly larger one, AG News, better suited for deep learning methods. \textbf{1)~20Newsgroups}~\footnote{\url{http://qwone.com/~jason/20Newsgroups/}}: We only take the articles from six top-level classes: \emph{computer, recreation, science, miscellaneous, politics, religion}, like in \cite{acl2019}. This dataset is relatively small, but a classic for NLP tasks (for each class, there are between 577-2856 samples for training and 382-1909 for validation). \textbf{2)~AG News}~\cite{ag_news}: This topic classification corpus was gathered from multiple news sources, for over more than one year \footnote{\url{http://groups.di.unipi.it/~gulli/AG_corpus_of_news_articles.html}}. It contains four topics, each class with 30000 samples for training and 1900 for validation.

\paragraph{Model and Training.} For training the DATE network we follow the pipeline in Fig.~\ref{fig:date_train}. In addition to the parameterized generator, we also consider a \emph{random generator}, in which we replace the masked tokens with samples from a uniform distribution over the vocabulary.
The \emph{discriminator} is composed of four Transformer layers, with \emph{two prediction heads} on top (for RMD and RTD tasks). We provide more details about the model in the supplementary material~\ref{apx:model_implementation_details}. We train the networks with AdamW with amsgrad~\cite{adamw-amsgrad}, $1e^{-5}$ learning rate, using sequences of maximum length $128$ for AG News, and $498$ for 20Newsgroups. We use $K=50$ predefined masks, covering $50\%$ of the input for AG News and $K=25$, covering $25\%$ for 20Newsgroups. The training converges on average after $5000$ update steps and the inference time is $0.005$ sec/sample in PyTorch~\cite{pytorch}, on a single GTX Titan X.


\subsection{Ablation studies}
\label{sec: ex-ablation}
To better understand the impact of different components in our model and making the best decisions towards a higher performance, we perform an extensive set of experiments (see Tab.~\ref{tab: ablation}). Note that we successively treat each AG News split as inlier and report the mean and standard deviations over the four splits. The results show that our model is robust to domain shifts.

\setlength{\tabcolsep}{3pt}
\begin{table}[t!]
\begin{center}
	\begin{tabular}{r r l c}
		\toprule
		\multicolumn{1}{p{0.1cm}}{Abl.} &
        \multicolumn{1}{p{1.5cm}}{\raggedright \;\;\;\;\;\;\;\;\;\;\;Method} &
        \multicolumn{1}{p{0.5cm}}{\raggedleft Variation} &
        \multicolumn{1}{p{0.9cm}}{\centering AUROC(\%)} \\
        \midrule
        & CVDD  & best & 83.1 $\pm$ \small 4.4 \\
        
        & OCSVM & best & 84.0 $\pm$ \small 5.0 \\ 
        
        & ELECTRA & adapted for AD & 84.6  $\pm$ \small 4.5\\ 
        
        & \textbf{DATE} & \textbf{(Ours)} & \textbf{90.0} $\pm$ \small 4.2 \\ 
        \midrule
        \midrule
        A. & Anomaly score  & MP &  72.4 $\pm$ \small  3.7\\
        &         		    & NE &  73.1 $\pm$ \small  3.9\\
		\midrule
		B. &  Generator   & small & 89.3  $\pm$ \small 4.2 \\ 
		&                 & large   & 89.8  $\pm$ \small 4.4 \\
		\midrule
		C. & Loss func    & RTD only & 89.4 $\pm$ \small 4.4 \\
		&                 & RMD only & 85.9 $\pm$ \small 4.1 \\
        \midrule
		D. & Masking       & 5 masks  &  87.5 $\pm$ \small 4.5 \\
        & patterns         & 10 masks &  89.2 $\pm$ \small 4.3 \\
        &                  & 25 masks &  89.8 $\pm$ \small 4.3 \\
	    &                   & 100 masks & 89.8 $\pm$ \small 4.3 \\ %
        \midrule
        E. & Mask percent   & 15\% & 89.5 $\pm$ \small 4.1 \\ 
        &                   & 25\% & 89.5 $\pm$ \small 4.1 \\ 
		\bottomrule
    \end{tabular}
\end{center}
\caption{Ablation study. We show results for the competition and report ablation experiments which are only one change away from our best \textbf{DATE} configuration: \textbf{A}. $PL_{RTD}$; \textbf{B}. Rand \textbf{C}. RTD + RMD; \textbf{D}. 50 masks; \textbf{E}. 50\%. For the ELECTRA line, we use: A. $PL_{RTD}$; B. Rand; C. RTD only; D. Unlimited; E. 15\%.
A. The Anomaly Score used over classification probabilities shows that $PL_{RTD}$ (used in DATE) is the best in predicting anomalies, meaning that our self-supervised classification task is well defined, with few ambiguous samples; B. A learned Generator does not justify its training cost; C. RMD Loss proved to be complementary with RTD Loss, their combination (in DATE) increasing the score and stabilizes the training; D+E.}
\label{tab: ablation}
\end{table}

\noindent\textbf{A. Anomaly score.} We explore three anomaly scores introduced in the \emph{$E^{3}Outlier$} framework~\cite{neurips2019} on semi-supervised and unsupervised AD tasks in Computer Vision: Maximum Probability (MP), Negative Entropy (NE) and our modified Pseudo Label ($PL_{RTD}$). These scores are computed using the softmax probabilities from the final classification layer of the discriminator. PL is an ideal score if the self-supervised task manages to build and learn well separated classes. The way we formulate our mask prediction task enables a very good class separation, as theoretically proved in detail in the supplementary material~\ref{apx:disjoint_patterns}. Therefore, $PL_{RTD}$ proves to be significantly better in detecting the anomalies compared with MP and NE metrics, which try to compensate for ambiguous samples.

\noindent\textbf{B. Generator performance.} We tested the importance of having a learned generator, by using a one-layer Transformer with hidden size 16 (small) or 64 (large). The \emph{random generator} proved to be better than both parameterized generators.

\noindent\textbf{C. Loss function.} For the final loss, we combined RTD (which sanctions the prediction per token) with our RMD (which enforces the detection of the mask applied on the entire sequence). We also train our model with RTD or RMD only, obtaining weaker results. This proves that combining losses with supervisions at different scales (locally: token-level and globally: sequence-level) improves AD performance. Moreover, when using only the RTD loss, the training can be very unstable (AUROC score peaks in the early stages, followed by a steep decrease). With the combined loss, the AUROC is only stationary or increases with time.

\noindent\textbf{D. Masking patterns.} The mask patterns are the root of our task formulation, hiding a part of the input tokens and asking the discriminator to classify them. As experimentally shown, having more mask patterns is better, encouraging increased expressiveness in the embeddings. Too many masks on the other hand can make the task too difficult for the discriminator and our ablation shows that having more masks does not add any benefit after a point. We validate the percentage of masked tokens in \textbf{E. Mask percent} ablation.

\subsection{Comparison with other AD methods}
\label{sec: ex-ssad}
We compare our method against classical AD baselines like Isolation Forest~\cite{iso_forest} and existing state-of-the-art OneClassSVMs~\cite{ocsmv} and CVDD~\cite{acl2019}. We outperform all previously reported performances on all \emph{20Newsgroups} splits by a large margin: 13.5\% over the best reported CVDD and 11.7\% over the best OCSVM, as shown in Tab.~\ref{tab: main_experiment}. In contrast, DATE uses the same set of hyper-parameters for a dataset, for all splits. For a proper comparison, we keep the same experimental setup as the one introduced in \cite{acl2019}.

\renewcommand{\thefootnote}{\fnsymbol{footnote}}
\setlength{\tabcolsep}{4pt}
\begin{table}[t]
\begin{center}
	\begin{tabular}{l l r  r  r  r }
		\toprule
		\multicolumn{1}{p{0.1cm}}{}&
        \multicolumn{1}{p{0.7cm}}{\raggedleft Inlier\\class} &
        \multicolumn{1}{p{1.2cm}}{\raggedleft IsoForest\\best} &
        \multicolumn{1}{p{1.2cm}}{\raggedleft OCSVM\\best} &
        \multicolumn{1}{p{1cm}}{\raggedleft CVDD\\best} &
        \multicolumn{1}{p{1cm}}{\raggedleft \textbf{DATE (Ours)}} \\
        \midrule
        \parbox[t]{2mm}{\multirow{6}{*}{\rotatebox[origin=c]{90}{\textbf{20 News}}}} & 
        comp & 66.1 & 78.0 & 74.0& \textbf{92.1}\\
        & rec & 59.4 & 70.0 & 60.6& \textbf{83.4}\\
        & sci & 57.8 & 64.2 & 58.2& \textbf{69.7}\\
        & misc & 62.4 &62.1& 75.7& \textbf{86.0}\\
        &pol & 65.3 & 76.1 & 71.5& \textbf{81.9}\\
        &rel & 71.4 & 78.9& 78.1& \textbf{86.1}\\
        \midrule        
        \parbox[t]{2mm}{\multirow{4}{*}{\rotatebox[origin=c]{90}{\textbf{AG News}}}} & 
        business & 79.6& 79.9 & 84.0\footnotemark[\value{footnote}] & \textbf{90.0}\\
        & sci & 76.9 & 80.7 & 79.0\footnotemark[\value{footnote}] & \textbf{84.0}\\
        & sports &  84.7& 92.4 & 89.9\footnotemark[\value{footnote}] & \textbf{95.9}\\
        & world & 73.2 & 83.2 & 79.6\footnotemark[\value{footnote}] & \textbf{90.1}\\
		\bottomrule
    \end{tabular}
\end{center}
\caption{Semi-supervised performance (AUROC\%). We test on the 20Newsgroups and AG News datasets, by comparing DATE against several strong baselines and state-of-the-art solutions (with multiple variations, choosing the best score per split as detailed in Sec.~\ref{sec: ex-ssad}): IsoForest, OCSVM, and CVDD. We largely outperform competitors with an average improvement of 13.5\% on 20Newsgroups and 6.9\% on AG News compared with the next best solution. Note that DATE uses the same set of hyper-parameters per dataset.}
\label{tab: main_experiment}
\end{table}

\paragraph{Isolation Forest.} We apply it over fastText or Glove embeddings, varying the number of estimators $(64, 100, 128, 256)$, and choosing the best model per split. In the unsupervised AD setup, we manually set the percent of outliers in the train set.

\paragraph{OCSVM.} We use the One-Class SVM model implemented in the CVDD work~\footnotemark[\value{footnote}]. For each split, we choose the best configuration (fastText vs Glove, rbf vs linear kernel, $\nu$ $\in$ [0.05, 0.1, 0.2, 0.5]).

\paragraph{CVDD.} This model~\cite{acl2019} is the current state-of-the-art solution for AD on text. For each split, we chose the best column out of all reported context sizes ($r$). The scores reported using the $c^*$ context vector depends on the ground truth and it only reveals "the potential of contextual anomaly detection", as the authors mention.

\footnotetext{Experiments done using the CVDD published code \url{https://github.com/lukasruff/CVDD-PyTorch}.}

\subsection{Unsupervised AD}
\label{sec: ex-uad}
\begin{figure}[t!]
\centering
\includegraphics[width=0.99\columnwidth]{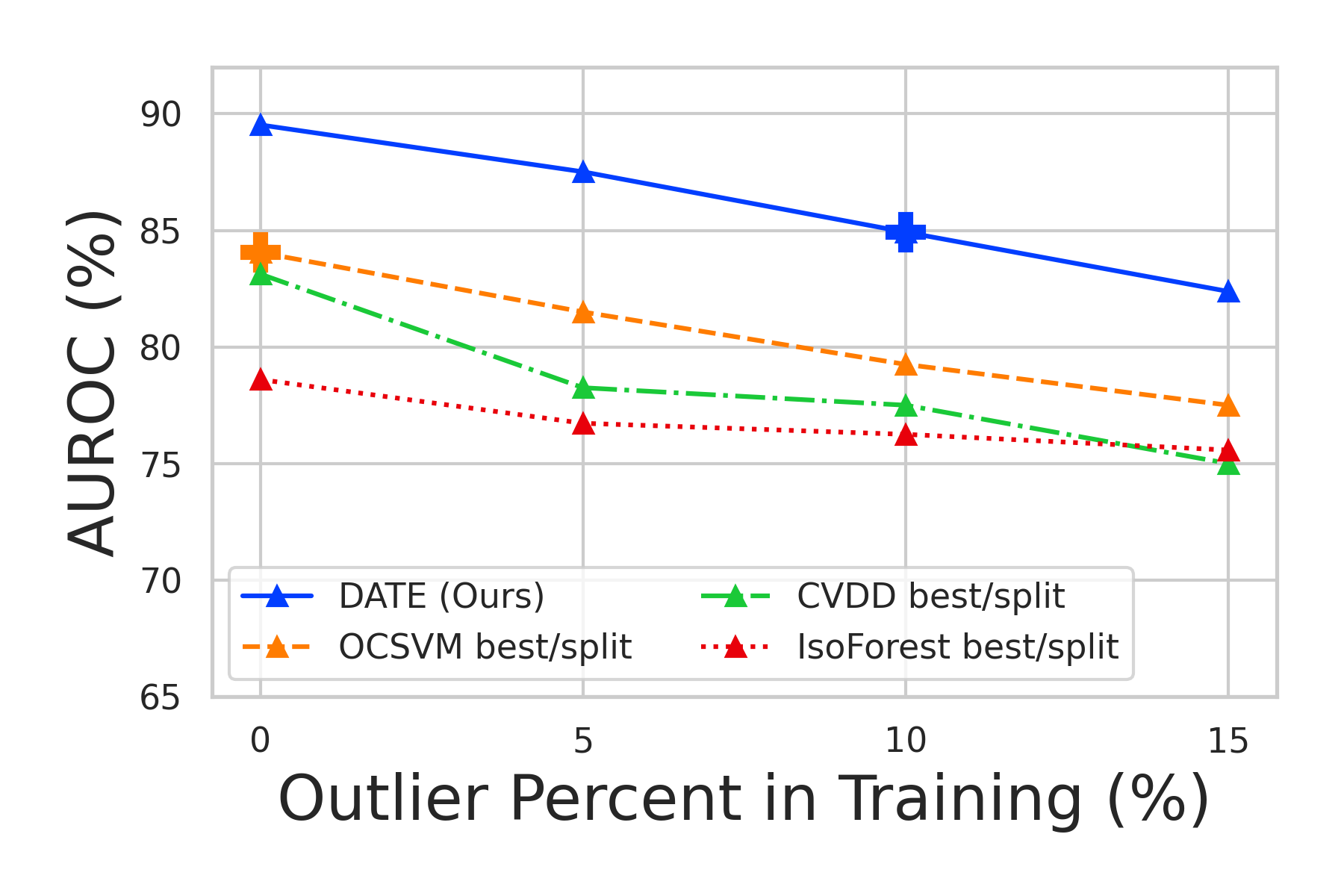}
\caption{Unsupervised AD. We test the performance of our method when training on impure data, which contains anomalies in various percentages: 0\%-15\%. The performance slowly decreases when we increase the anomaly percentage, but even at 10\% contamination, it is still better than state-of-the-art results on self-supervised anomaly detection in text \cite{acl2019}, which trains on 0\% anomalous data, proving the robustness of our method. Experiments were done on all AG News splits.}
\label{fig: fully_unsup_od}
\end{figure}

We further analyse how our algorithm works in a fully unsupervised scenario, namely when the training set contains some anomalous samples (which we treat as normal ones). By definition, the quantity of anomalous events in the training set is significantly lower than the normal ones. In this experiment, we show how our algorithm performance is influenced by the percentage of anomalies in training data. Our method proves to be extremely robust, surpassing state-of-the-art, which is a semi-supervised solution, trained over a clean dataset (with 0\% anomalies), even at 10\% contamination, with +0.9\% in AUROC (see Fig.~\ref{fig: fully_unsup_od}). By achieving an outstanding performance in the unsupervised setting, we make unsupervised AD in text competitive against other semi-supervised methods. The reported scores are the mean over all AG News splits. We compare against the same methods presented in Sec.~\ref{sec: ex-ssad}.

\subsection{Qualitative results}
\begin{figure}[t!]
\centering
\includegraphics[width=0.99\columnwidth]{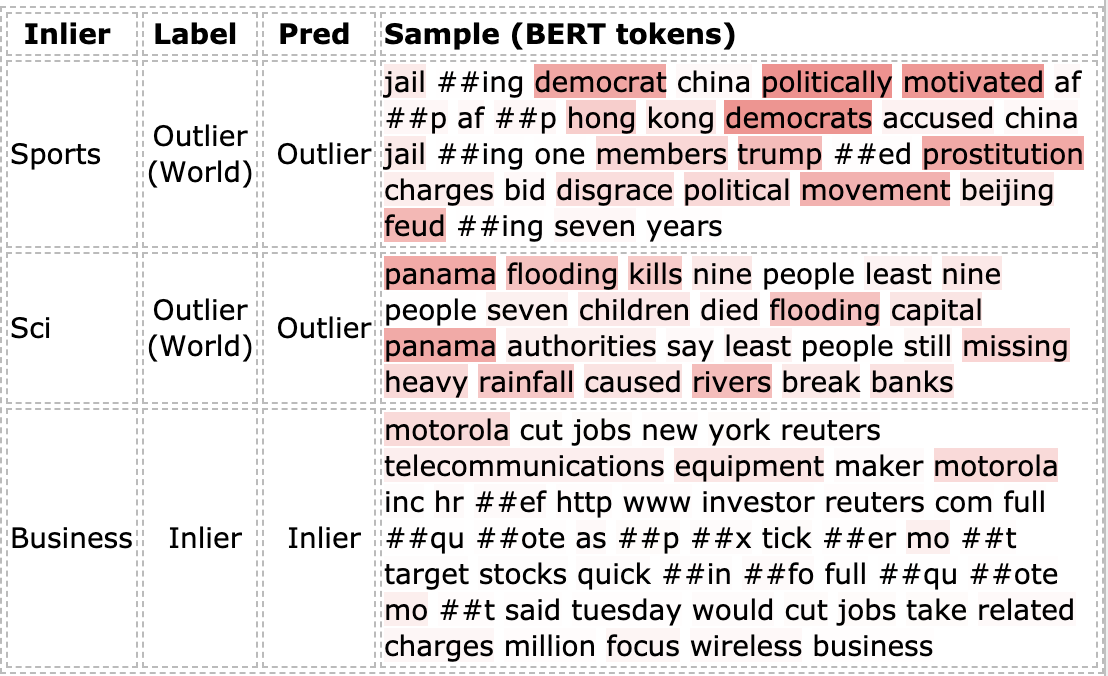} 
\caption{Qualitative examples. Lower scores are shown in a more intense red, and point to anomalies. In the $1^{st}$ example, words from politics are flagged as anomalous for sports. In the $2^{nd}$ one, words describing natural events are outliers for technology. In the $3^{rd}$ row, while few words have higher anomaly potential for the business domain, most of them are appropriate.}
\label{fig: qual_ex}
\end{figure}

We show in Fig.~\ref{fig: qual_ex} how DATE performs in identifying anomalies in several examples. Each token is colored based on its PL score.

\begin{figure}[t!]
\centering
    \includegraphics[width=0.49\columnwidth]{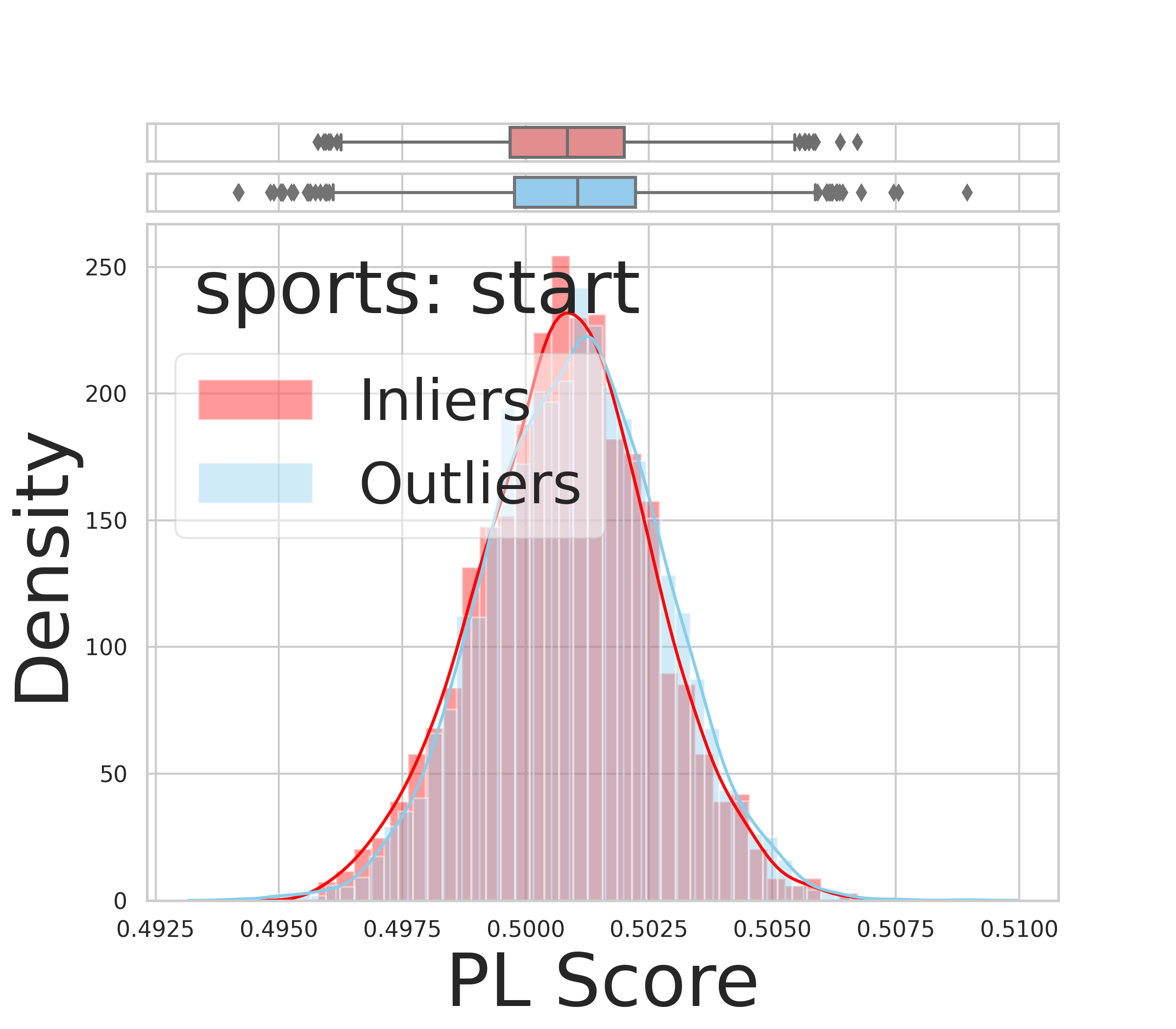}
    \includegraphics[width=0.49\columnwidth]{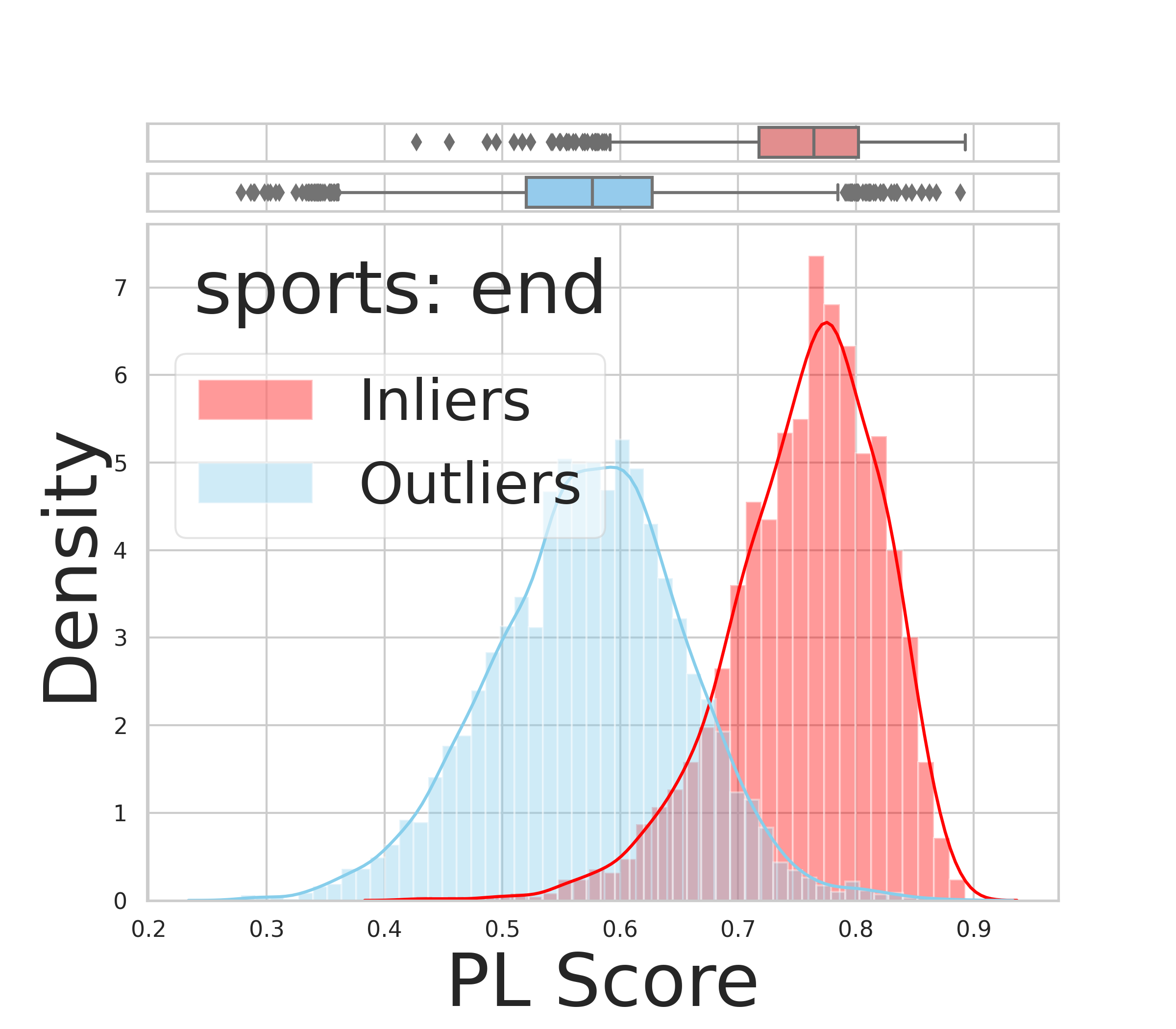} 
    \includegraphics[width=0.49\columnwidth]{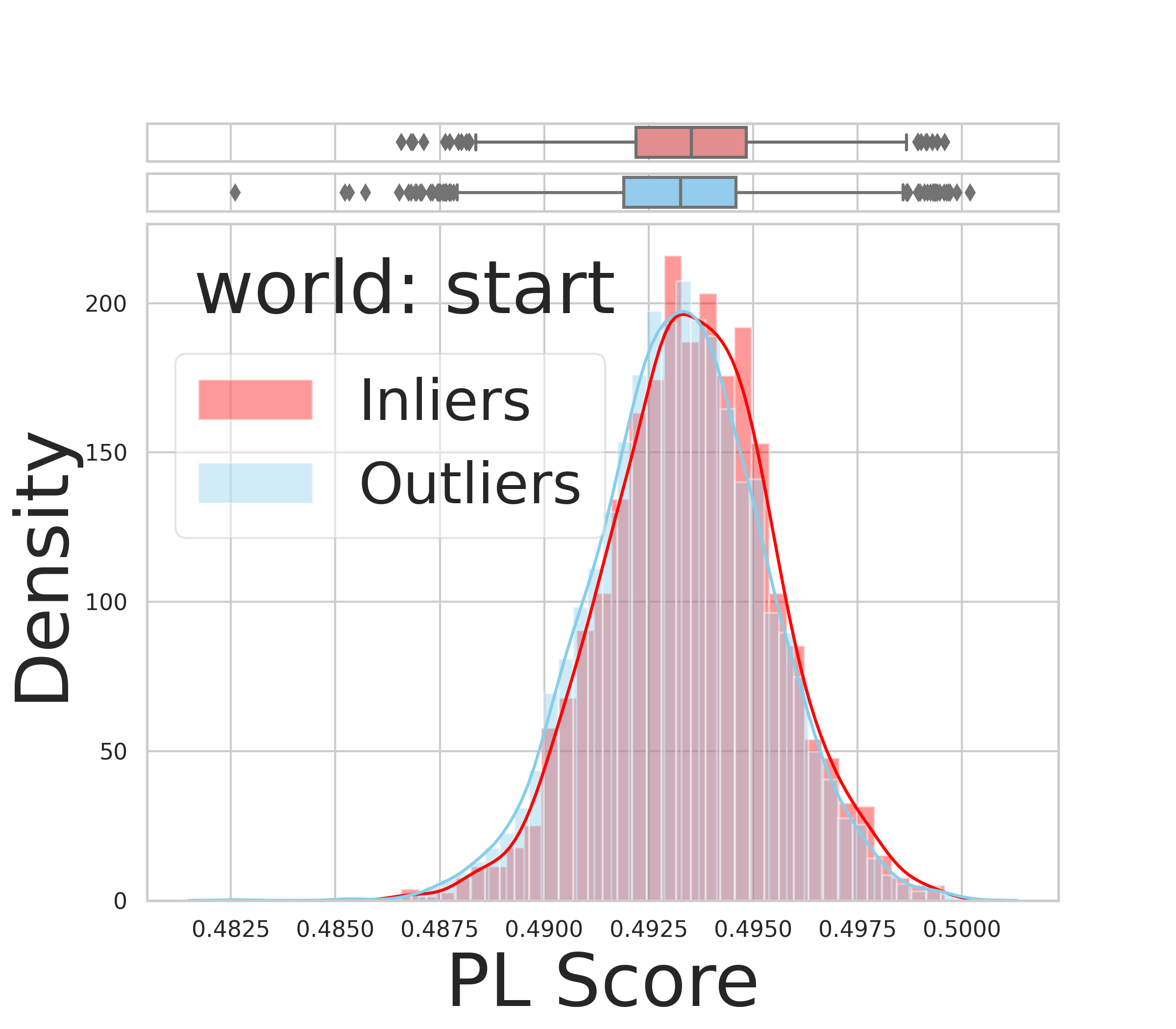}
    \includegraphics[width=0.49\columnwidth]{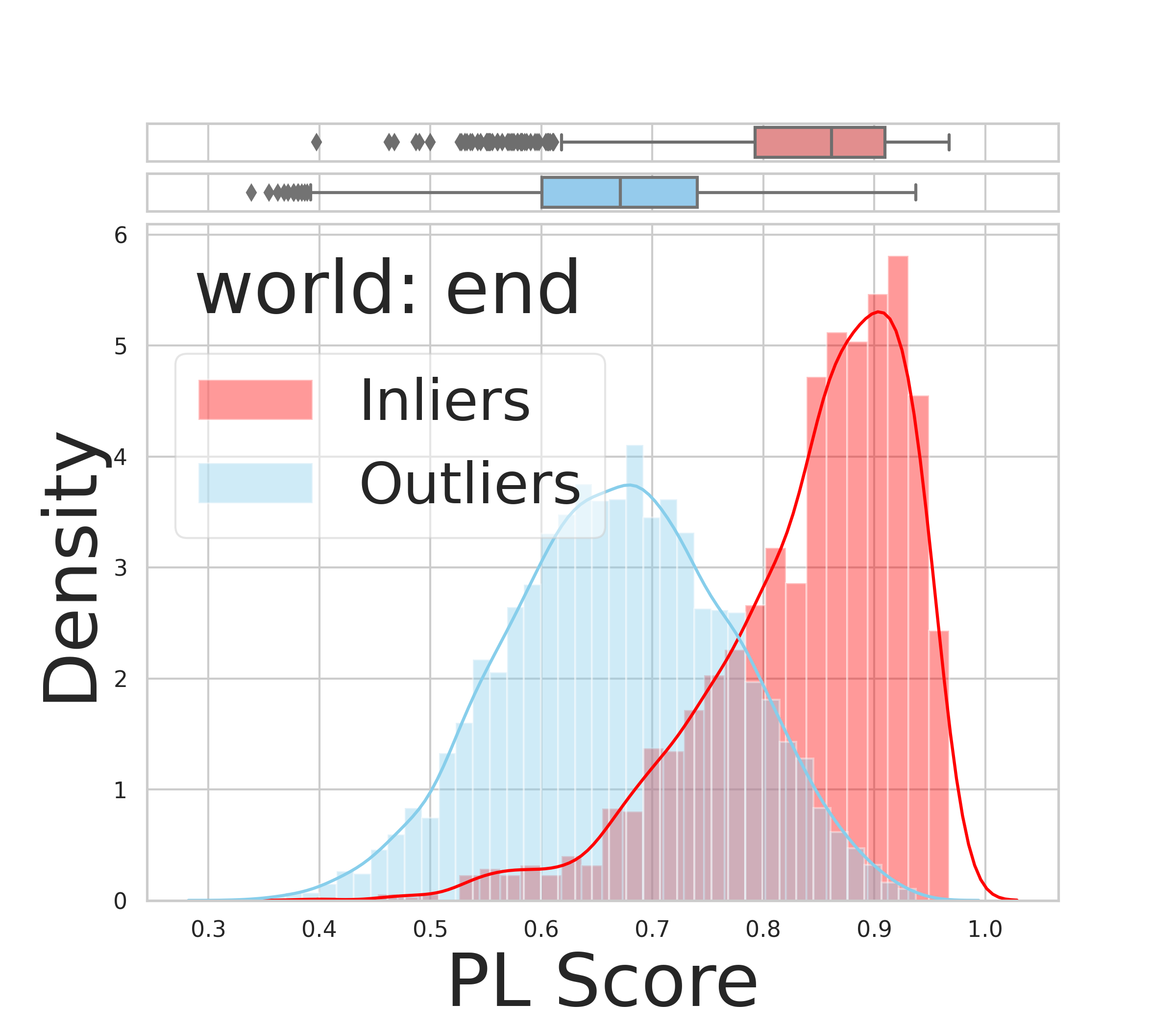}
\caption{Normalized histogram for anomaly score. We see how the anomaly score (PL) distribution varies among inliers and outliers, from the beginning of the training ($1^{st}$ column) to the end ($2^{nd}$ column), where the two become well separated, with relatively low interference between classes. Note that a better separation is correlated with high performance ($1^{st}$ line split has $95.9\%$ AUROC, while the $2^{nd}$ has only $90.1\%$).}
\label{fig: histogram}
\end{figure}

\paragraph{Separating anomalies.} We show how our anomaly score (PL) is distributed among normal vs abnormal samples. For visualization, we chose two splits from AG News and report the scores from the beginning of the training to the end. We see in Fig.~\ref{fig: histogram} that, even though at the beginning, the outliers' distribution of scores fully overlaps with the inliers, at the end of training the two are well separated, proving the effectiveness of our method.

\section{Conclusion}
We propose DATE, a model for tackling Anomaly Detection in Text, and formulate an innovative self-supervised task, based on masking parts of the initial input and predicting which mask pattern was used. After masking, a generator reconstructs the initially masked tokens and the discriminator predicts which mask was used.
We optimize a loss composed of both token and sequence-level parts, taking advantage of powerful supervision, coming from two independent pathways, which stabilizes learning and improves AD performance.
For computing the anomaly score, we alleviate the burden of aggregating predictions from multiple transformations by introducing an efficient variant of the Pseudo Label score, which is applied per token, only on the original input.
We show that this score separates very well the abnormal entries from normal ones, leading DATE to outperform state-of-the-art results on all AD splits from 20Newsgroups and AG News datasets, by a large margin, both in the semi-supervised and unsupervised AD settings.

\paragraph{Acknowledgments.} This work has been supported in part by UEFISCDI, under Project PN-III-P2-2.1-PTE-2019-0532.

\clearpage

\bibliographystyle{acl_natbib}
\bibliography{naacl}

\appendix

\section{Disjoint patterns analysis}
\label{apx:disjoint_patterns}
We start from two observations regarding the performance of DATE, our Anomaly Detection algorithm. First, a discriminative task performs better if the classes are well separated \cite{Deng2012} and there is a low probability for confusions. Second, the PL score for anomalies achieves best performance when the probability distribution for its input is clearly separated. Intuitively, for three classes, PL([0.9, 0.05, 0.05]) is better than PL([0.5, 0.3, 0.2]) because it allows PL to give either near $1$ score if the class is correct, either near $0$ score if it is not, avoiding the zone in the middle where we depend on a well chosen threshold.

Since the separation between the mask patterns greatly influences our final performance, we next analyze our AD task from the mask pattern generation point of view. Ideally, we want to have a sense of how disjoint our randomly sampled patterns are and make an informed choice for the pattern generation hyper-parameters.

First, we start by computing an upper bound for the probability of having two patterns with at least $p$ common masked points. We have $\binom{S}{M}$ patterns, where $S$ is the sequence length and $M$ is the number of masked tokens. We fix the first $p$ positions that we want to mask in any pattern. Considering those fixed masks, the probability of having a sequence with $M$ masked tokens, with $p$ tokens in the first positions is $r$:

\begin{equation}
\label{eq: r_ratio}
    r = \frac{\binom{S-p}{M-p}}{\binom{S}{M}}.
\end{equation}

Next, the probability that two sequences mask the first $p$ tokens is $r^2$. But we can choose those two positions in a $\binom{S}{p}$ ways. So the probability that any two sequences have at least $p$ common masked tokens is lower than $UB_{2}$:

\begin{equation}
\label{eq: concide_bound}
    UB_{2} = \binom{S}{p} r^2\\
\end{equation}

Next, out of our generated patterns, we sample $N$ masks, so the probability becomes less than the upper bound $UB_{N}$:
\begin{equation}
\label{eq: final_bound}
    \begin{aligned}
    UB_{N} &= \binom{N}{2} UB_{2} = \binom{N}{2} \binom{S}{p} r^2 \\ 
        &= \binom{N}{2} \binom{S}{p} (\frac{\binom{S-p}{M-p}}{\binom{S}{M}})^2.
    \end{aligned}
\end{equation}

In our experiments, the sequence length is $S = 128$ and we chose the number of masked tokens to be between 15\% and 50\% ($M$ between 19 and 64). We consider that two patterns are disjoint when they have less than $p$ masked tokens in common, for $N$ sampled patterns.

The probability that any two patterns collide (have more than $p$ masked tokens in common) is very low. We compute several values for its upper bound: $UB_{N=100, p=12}=5e-4$, $UB_{N=100, p=15}=1e-9$, $UB_{N=10, p=15}=1e-11$, $UB_{N=10, p=13}=1e-7$.

In conclusion, for our specific setup, the probability for two masks to largely overlap (large $p$ compared with $S$) is extremely small, ensuring us a good performance in the discriminator. We take advantage of this property of our pretext task by combining the discriminator output probabilities with the PL score.

\section{Model implementation}
\label{apx:model_implementation_details}
We add next more details on the implementation of the modules: from the ablation experiments in Tab.~\ref{tab: ablation}, \textbf{Generator (small)}: 1 Transformer layer, with 4 self-attention heads, token and positional embeddings of size 128, hidden layer of size 16, feedforward layer of sizes 1024 and 16; \textbf{Generator (large)}: 1 Transformer layer, with 4 self-attention heads, token and positional embeddings of size 128, hidden layer of size 64, feedforward layer of sizes 1024 and 64;
As empirical experiments showed us, we choose a \emph{random Generator} (samples were drawn from a uniform distribution over the vocabulary) in our final model.
\textbf{Discriminator}: 4 Transformer layers, each with 4 self-attention heads, hidden layers of size 256, feedforward layers of sizes of 1024 and 256, 128-dimensional token and positional embeddings, which are \emph{tied} with the generator. For other unspecified hyper-parameters we use the ones in ELECTRA-Small model.
\textbf{Prediction Heads}: both heads have 2 linear layers separated by a non-linearity, ending in a classification. \textbf{Loss weights:} We set the RTD $\lambda$ weight to 50 as in \cite{Clark2020}, and the RMD $\mu$ weight to 100.

\section{More qualitative and quantitative Results}
\label{apx:prauc_experiments}
\begin{figure}[t]
\centering
\includegraphics[width=0.99\columnwidth]{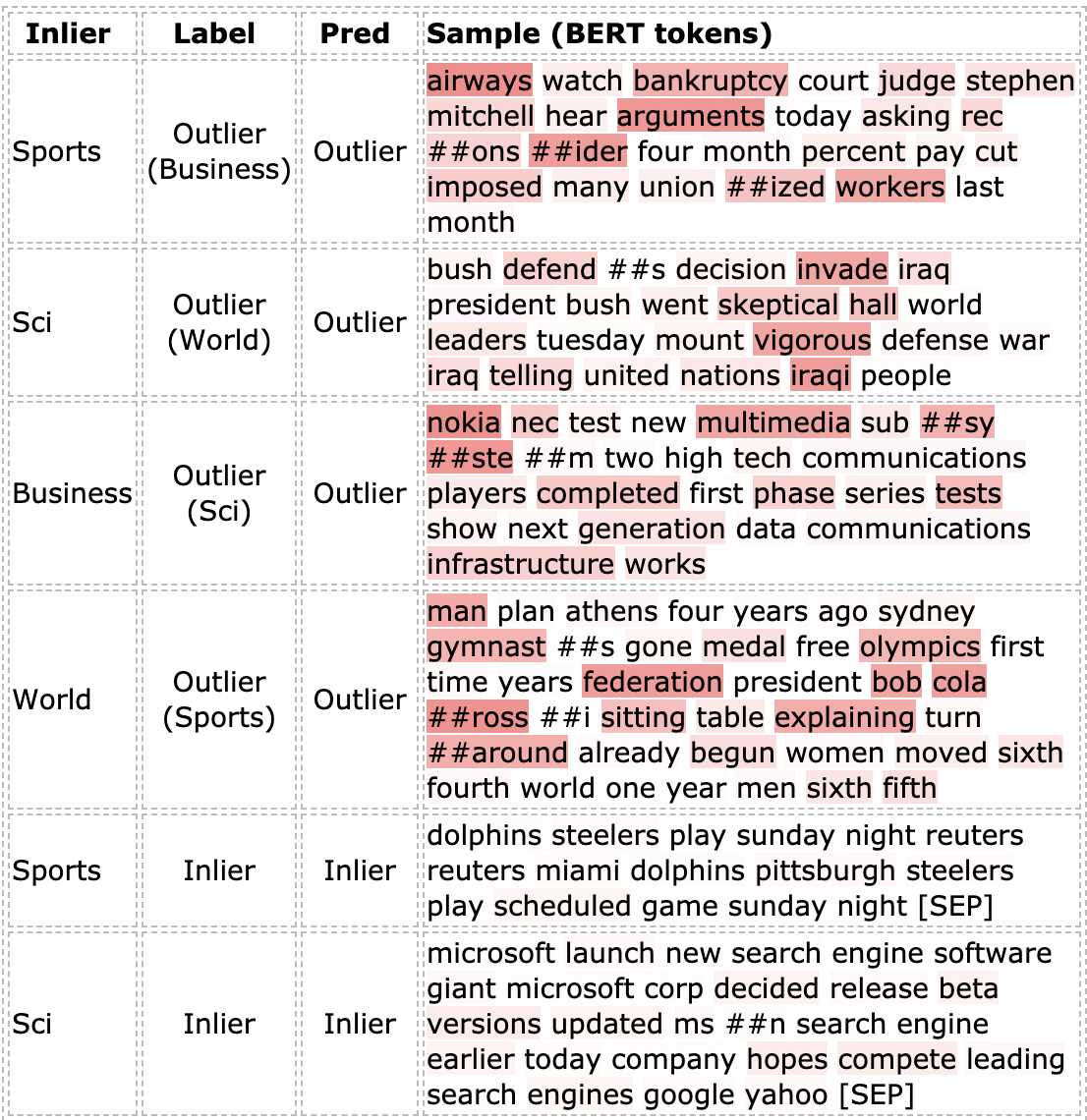} 
\caption{More qualitative examples.}
\label{fig: qual_ex_2}
\end{figure}

In Fig.~\ref{fig: qual_ex_2} we show more qualitative results, trained on different inliers. To encourage further more detailed comparisons, we report the AUPR metric on AG News for inliers and outliers (see Tab.~\ref{tab:prauc_experiments}). When all the other metrics are almost saturated, we notice that AUPR-in better captures the performance on a certain split.

\begin{table}[t]
\centering
\begin{tabular}{l  r r r r}
\toprule
\multicolumn{1}{p{1cm}}{Subset} & business & sci & sports & world\\
\midrule
AUPR-in  & 74.8 & 62.4 & 88.8 & 81.9 \\
AUPR-out & 96.1 & 93.5 & 98.5 & 95.5 \\

\bottomrule
\end{tabular}
\caption{We report AUPR metric for AG News splits, on inliers and outliers since this is a more relevant metric for unbalanced classes (which is the case for all splits in text AD, as explained in Anomalies setup).}
\label{tab:prauc_experiments}
\end{table}

\end{document}